\documentclass{article}

\usepackage[final]{corl_2022} 

\usepackage{etoolbox}

\usepackage[utf8]{inputenc} 
\usepackage[T1]{fontenc}    
\usepackage{natbib}
\usepackage{url}            
\usepackage{booktabs}       
\usepackage{amsfonts}       
\usepackage{nicefrac}       
\usepackage{microtype}      
\usepackage{paralist}       
\usepackage{graphicx}
\usepackage{amsmath}
\usepackage[ruled,vlined,linesnumbered]{algorithm2e}
\usepackage{bbm}

\makeatletter
\patchcmd\algocf@Vline{\vrule}{\vrule \kern-0.4pt}{}{}
\patchcmd\algocf@Vsline{\vrule}{\vrule \kern-0.4pt}{}{}
\makeatother

\usepackage[labelfont=bf]{caption}
\usepackage{subcaption}
\usepackage{diagbox}
\usepackage{bbding}
\usepackage{booktabs}
\usepackage{pifont}
\usepackage{wrapfig}
\usepackage{lipsum}
\usepackage{multirow}
\usepackage{enumitem}
\usepackage[font={footnotesize}]{caption}

\DeclareMathOperator*{\argmin}{arg\,min}

\newcommand{\rebuttal}[1]{{\color{black} #1}}
\newcommand{\algo}{PASTA}
\newcommand\blfootnote[1]{%
  \begingroup
  \renewcommand\thefootnote{}\footnote{#1}%
  \addtocounter{footnote}{-1}%
  \endgroup
}

\title{Planning with Spatial and Temporal Abstraction from Point Clouds for Deformable Object Manipulation}

\author{
    Xingyu Lin $\ast\dagger$ \\
    \And
    Carl Qi $\ast\dagger$ \\
    \And
    Yunchu Zhang $\dagger$ \\
    \And
    Zhiao Huang $\ddagger$ \\
    \AND
    Katerina Fragkiadaki $\dagger$ \\
    \And
    Yunzhu Li $\mathsection$ \\
    \And
    Chuang Gan $\mathparagraph$ \\
    \And
    David Held $\dagger$ \\
}

\begin{document}
\maketitle

\blfootnote{$\ast$ equal contribution; $\dagger$ CMU $\ddagger$ UC San Diego; $\mathsection$ Stanford University; $\mathparagraph$ UMass Amherst \& MIT-IBM Lab}
\begin{abstract}
Effective planning of long-horizon deformable object manipulation requires suitable abstractions at both the spatial and temporal levels. Previous methods typically either focus on short-horizon tasks or make the strong assumption that full-state information is available. \rebuttal{However, full states of deformable objects are often unavailable.} In this paper, we propose PlAnning with \rebuttal{Spatial and Temporal Abstraction} (PASTA), which incorporates both spatial abstraction (reasoning about objects and their relations to each other) and temporal abstraction (reasoning over skills instead of low-level actions). Our framework maps high-dimension \rebuttal{3D point clouds} into a set of latent vectors and plans skill sequences with the latent set representation. Our method can solve challenging, novel sequential deformable object manipulation tasks in the real world, which require combining multiple tool-use skills such as cutting with a knife, pushing with a pusher, and spreading dough with a roller. Additional materials can be found on our project website.\footnote{\url{https://sites.google.com/view/pasta-plan}}
\end{abstract}



\begin{figure}[ht]
    \centering
    \includegraphics[width=\textwidth]{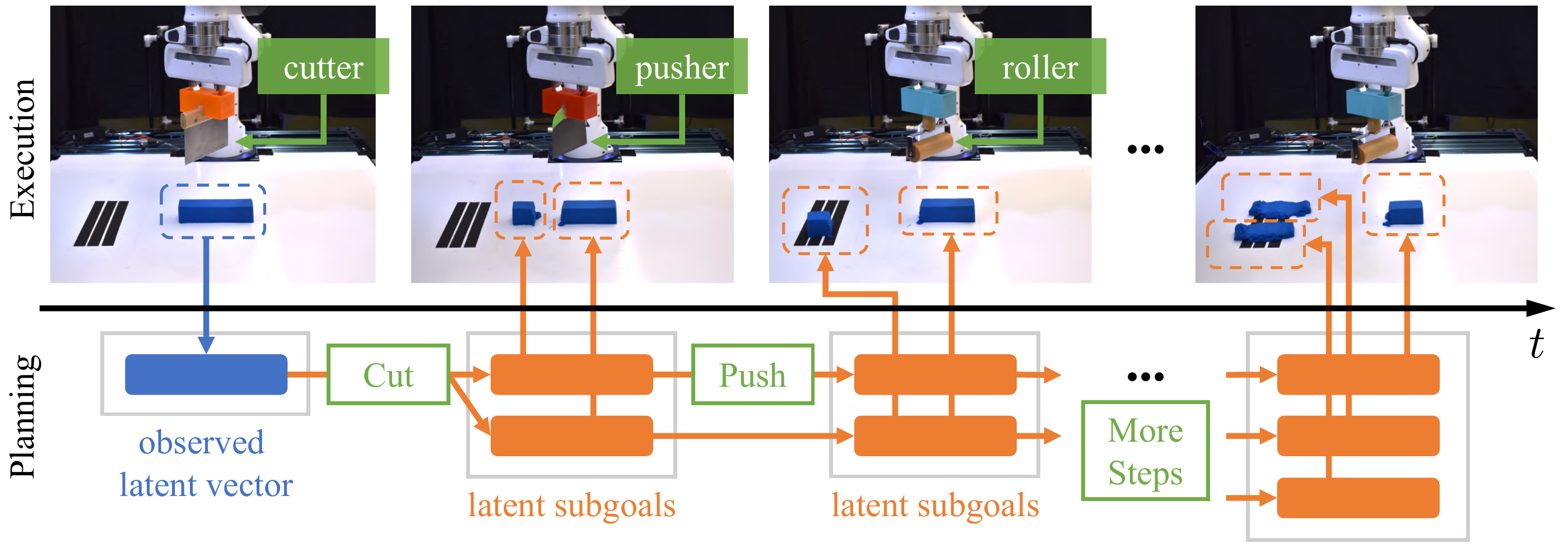}
    \caption{\textbf{Long-horizon dough manipulation with diverse tools.}
    Our framework is able to solve  long-horizon, multi-tool, deformable object manipulation tasks that the agent has not seen during training. The illustrated task here is to cut a piece of dough into two with a cutter, transport the pieces to the spreading area on the left (with a high-friction surface) using a pusher, and then flatten both pieces with a roller.}
    \label{fig:pull}
\end{figure}

\section{Introduction}

Consider a typical cooking task of making dumplings from dough.  People plan over which piece of dough to manipulate and which tool to use in sequence, incorporating both spatial and temporal abstractions. A spatial abstraction reasons about objects, parts, and their relations to each other, such as reasoning about pieces of dough 
instead of reasoning about individual dough \rebuttal{particles}; such a spatial abstraction enables efficient planning and compositional generalization.
On the other hand, a temporal abstraction incorporates abstract actions represented as a set of skills. \rebuttal{With abstract actions, the agent plans for the types and parameters of the skills to execute over a period of time, instead of making plans for low-level actions such as joint torques at each time step}. Temporal abstractions allow planning at the skill level, enabling more efficient optimization for solving long-horizon tasks.
An autonomous robot that operates in unstructured environments should be able to reason about world dynamics using high-level spatial and temporal abstractions instead of reasoning only over the \rebuttal{physical state, raw sensory observation}, or low-level robot actions.



The research community is making rapid progress towards developing state abstractions for manipulating \rebuttal{deformable objects, including
key points~\cite{ma2022learning,manuelli2019kpam},} graphs~\cite{li2018learning,lin2021VCD}, dense object descriptors~\cite{florence2018dense}, or implicit functions~\cite{li20223d,driess2022learning}. However, most of these approaches do not make abstractions at the temporal level, limiting their use to short-horizon tasks. Methods are also being developed with temporal abstractions, planning over a set of skills to solve long-horizon tasks~\cite{lin2022diffskill,nasiriany2022maple,dalal2021accelerating}. However, the lack of spatial abstraction severely limits their generalization ability. Therefore, it remains a key question in robot learning on how to learn spatial and temporal abstractions within a unified framework for complex and long-horizon manipulation tasks. 

In this work, we focus on the challenging task of sequential deformable object manipulation, as shown in Figure~\ref{fig:pull}. We consider a set of dough manipulation tasks that require sequentially applying different skills using multiple tools to manipulate dough, such as spreading using a roller, cutting using a knife, and pushing using a pusher, where the longest task requires applying 6 skills in sequence. Deformable objects like dough have nearly infinite degrees of freedom. As such, in this work, we dynamically cluster points in a point cloud into different groups and learn a point cloud encoder to map each element in the group into a latent vector. In this way, we obtain a compositional 3D set representation of the state space. Given an observation and a target point cloud, we then sample skill sequences with subgoals generated in this latent space.  We learn skill abstraction modules to determine the feasibility and score of each skill sequence and use them for planning. 

Our contribution of this paper is a framework that PlAns with Spatial and Temporal Abstraction (PASTA) by learning a set of skill abstraction modules over a 3D set representation. Our framework can compose a set of skills to solve complex tasks with more entities and longer-horizon than what was seen during training. We show that PASTA significantly outperforms an ablation that performs planning with a flat representation without a spatial abstraction (e.g. without a set representation). Finally, our planner can be trained in simulation and transferred to the real world.

\section{Related Work}
\textbf{Model-based Planning for Sequential Manipulation.}
\rebuttal{One line of research for sequential manipulation is Task and Motion Planning (TAMP)}. TAMP systems typically assume known object states and known effects for the action operators~\cite{garrett2020pddlstream, toussaint2018differentiable,fikes1971strips, mcdermott1998pddl,  toussaint2015logic}. However, it is difficult to estimate states and dynamics for unknown objects or from partial observations. While recent works have made progress in learning certain components of the system, such as the logical states~\cite{yuan2022sornet} from high dimensional observations or learning action models~\cite{liang2021search, ugur2015bottomup, wang2021learning} from interactions, they still require either known states or known action operators. In contrast, we do not assume known states or action operators, and learn a 3D set representation as well as the action model with the representation.

Another approach learns dynamics directly from visual observations~\cite{li2018learning,ebert2018visual,hafner2019learning}. Most of these works focus on learning a one-step dynamics model for planning short-horizon tasks. A few works learn the dynamics model over a set of skills and use it for sequential manipulation of rigid objects~\cite{fang2020dynamics,simeonov2020long} or deformable objects~\cite{lin2022diffskill}. However, these works do not use an object-centric representation and thus cannot easily generalize to more complex scenes. In contrast, our framework unifies both temporal and spatial abstraction and can perform long-horizon manipulation for complex tasks with more objects than in previous work, as we will show.

\textbf{Planning with Spatial Abstraction.}
Prior works leverage spatial abstraction to facilitate solving tasks that involve complex dynamics and high-dimensional observations. These works either model a compositional system with Graph Neural Networks (GNN)~\cite{ li2018learning, lin2021VCD, driess2022learning, pfaff2020learning, ma2021learning} or learn policies directly from object-centric representations~\cite{heravi2022visuomotor, devin2018deep}. These works demonstrate compositional generalization, but they learn policies or one-step dynamics models for planning, which can be difficult for solving long-horizon tasks. In contrast, our framework connects temporally extended spatial abstractions with a feasibility predictor to plan over a longer time horizon. \rebuttal{Xu et al.~\cite{xu2019regression} learns a planner grounded on object-centric visual observation and can solve long-horizon tasks. However, its planner takes a symbolic goal and plans in a predefined symbolic domain to output symbolic subgoals. In contrast, our method does not require defining a symbolic planning domain.}

\textbf{Deformable Object Manipulation.}
Deformable objects have nearly infinite degrees of freedom and complex dynamics, making them very challenging to manipulate. 
Previous works have explored pouring liquid~\cite{li20223d,schenck2017visual, icra2022pouring}, rope manipulation~\cite{sundaresan2020learning, mitrano2021learning}, and cloth manipulation~\cite{maitin2010cloth,fabric_vsf_2021,lin2021VCD,Huang2022MEDOR,weng2021fabricflownet}. Other papers have also explored manipulating elastoplastic objects such as deforming them by grasping~\cite{li2018learning,shi2022robocraft}, rolling~\cite{qi2022ral, figueroa2016learning, matl2021deformable}, or cutting~\cite{heiden2021disect}. However, these works mostly only consider manipulation with one skill at a time. In contrast, we consider the task of sequential manipulation using multiple tools. The one exception is DiffSkill~\cite{lin2022diffskill}, where multiple skills are chained together. However, DiffSkill uses RGB-D images to represent the scene. \rebuttal{In contrast, we use a 3D set representation that separately encodes each entity in the scene, enabling compositional generalization to tasks with more objects and longer-horizon. Furthermore, we use point clouds as the input and we are able to transfer our planner from simulation to the real world.}


\begin{figure}
    \centering
    \includegraphics[width=\textwidth]{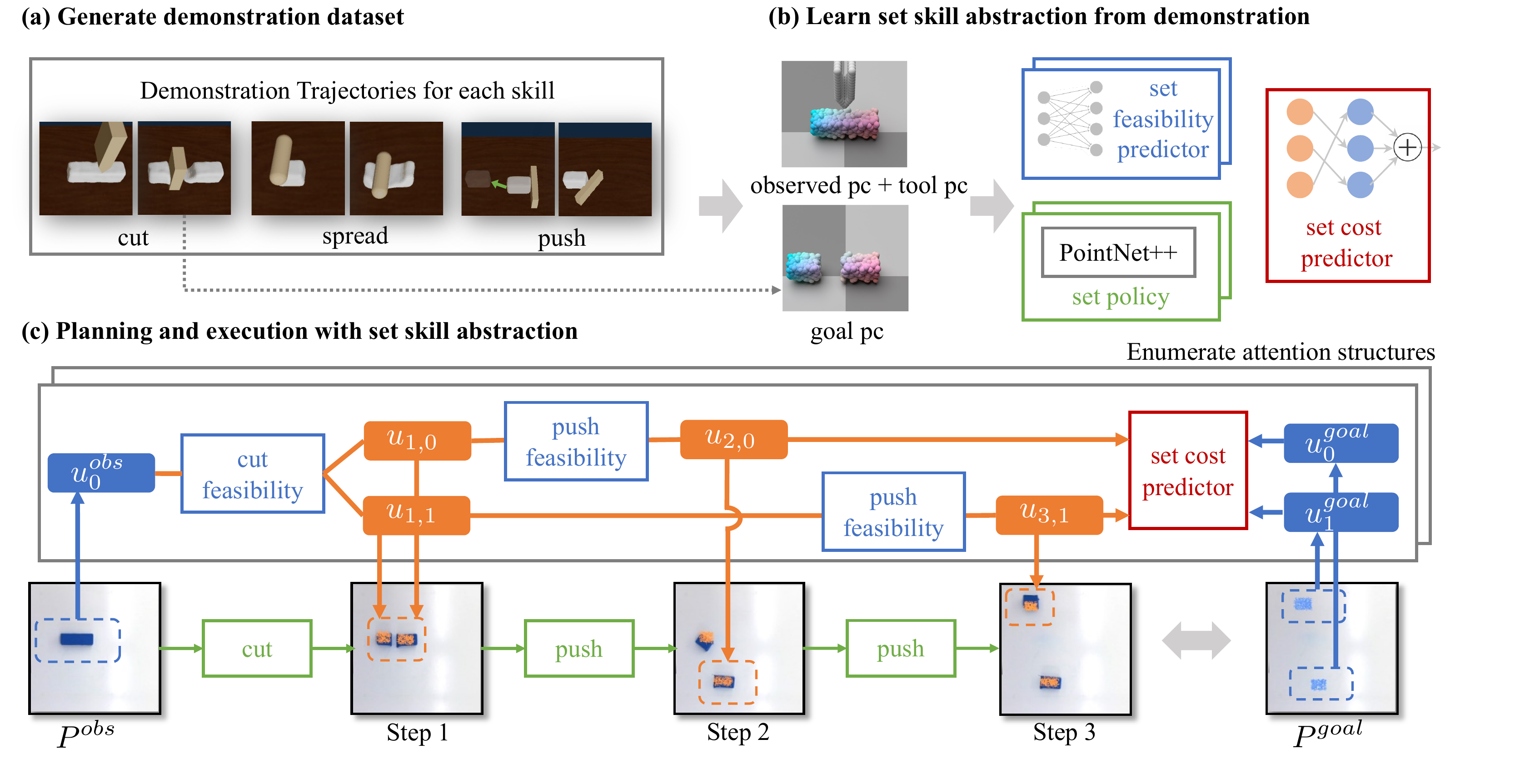}
    \caption{ \textbf{Overview of our proposed framework PASTA.} (a) We first generate demonstration trajectories for each skill in a differentiable simulator using different tools. (b) We then sample point clouds (pc) from the demonstration trajectories to train our set skill abstraction modules. (c) We map point clouds into a latent set representation and plan over tool-use skills to perform long-horizon deformable object manipulation tasks. $P^{obs}, P^{goal}$ are the observation and target pc; $u_{i,j}$ denotes component $j$ at step $i$. The example shows our method performs the CutRearrange task, which requires cutting the dough into two pieces with a knife and transporting each piece to its target location.} 
    \label{fig:overview}
\end{figure}

\section{Method}
Given a point cloud of the dough $P^{obs}$ and a goal point cloud $P^{goal}$, our objective is to execute a sequence of actions \rebuttal{$a_1, ..., a_{T_{tot}}$} that minimizes the distance between the final observed point cloud and the goal \rebuttal{$D(P^{obs}_{T_{tot}}, P^{goal})$ where $P^{obs}_{T_{tot}}$ is the observation point cloud at time $T_{tot}$}. We aim to solve long-horizon tasks that require chaining multiple skills in novel scenes with more objects than training. To do so, we present a general framework that incorporates spatial and temporal abstractions, as summarized in Figure~\ref{fig:overview}. We use point clouds as input to all our modules to enable easier transfer from simulation to the real world and to enable robustness to changes in viewpoint.

We assume access to an offline dataset of demonstration trajectories $\mathcal{D}_{demo}$ \rebuttal{from $K$ skills}, where each trajectory demonstrates \rebuttal{one of the skills} using one tool. We can learn skill policies by imitation learning from these demonstrations. To chain these skills to solve long-horizon tasks, we train a set of skill abstraction modules~(Sec.~\ref{sec:abstraction}) which can be used for planning in the latent space (Sec.~\ref{sec:latent-space}).

\subsection{Spatial Abstraction from Point Clouds}
\label{sec:latent-space}
\textbf{Scene Decomposition:} First, we describe our spatial abstraction of the point cloud observation. Given a point cloud $P\in \mathbb{R}^{N\times 3}$, we first cluster the points into different components based on their proximity in space. In this paper, we apply DBSCAN~\cite{ester1996density} to 
$P$ and group points into a set of entity point clouds $\{P_{i} \in \mathbb{R}^{N_i \times 3}\}_{i=1...C}$ by separating points from high-density regions into different clusters. While other works on scene decomposition can also be used~\cite{locatello2020object}, we find this simple method to be sufficient for our tasks.

\textbf{Entity Encoding:} 
Planning directly in the high-dimensional space of point clouds is inefficient. To enable efficient planning in a latent space, we  train a point cloud variational autoencoder (VAE)~\cite{pointflow}.
The VAE 
model includes three modules: A point cloud encoder $\phi: \mathbb{R}^{N_i\times 3} \rightarrow \mathcal{U}$ that maps \rebuttal{each entity point cloud} to a latent vector, a decoder $\psi:\mathcal{U} \rightarrow \mathbb{R}^{N_i\times 3}$ that maps from a latent space back into a point cloud, and a prior distribution over the latent space $p_u: \mathcal{U} \rightarrow [0, 1]$ which can be used to generate samples from the latent space during planning. \rebuttal{We can then encode the point clouds $\{P_i\}$ into a set of latent vectors: $\{u_i\}_{i=1...C}$ as our set latent representation.} We further achieve translation invariance by separating the translation from its shape embedding. See the Appendix for details.

\subsection{Learning Skills by Imitation}
Given the demonstration trajectories $\mathcal{D}_{demo}$ of $K$ skills, we first distill these trajectories into $K$ closed-loop policies. The input to the policy for the $k^{th}$ skill $\pi_k$ is a subset of the observed point clouds $\{P^{o}_i\}$ and goal point clouds $\{P^{g}_{j}\}$, and a tool point cloud $P^{tool}_{k}$. The policy only sees a subset of the dough point cloud and goal point cloud in the scene to enable compositional generalization to scenes with more objects. For example, a dough-spreading policy will only see the dough being spread. To achieve this, we train each set point cloud policy with behavior cloning and hindsight relabeling~\cite{andrychowicz2017hindsight} on the demonstration dataset with an attention mask that filters out the non-relevant entity point clouds. During planning, this attention mask will be provided by the planner. The policy outputs an action at each timestep to control the tool directly.

\subsection{Neural Spatial and Temporal Abstraction} \label{sec:abstraction}
We assume that each skill learned from the demonstration is only capable of performing \rebuttal{a single-stage task with a single tool for a single object. To solve longer-horizon tasks, we further learn a feasibility predictor and a \rebuttal{cost} predictor}. They can be used to plan subgoals that chain the skills into a sequence, such that each subgoal is feasible for the corresponding policy to reach and the final subgoal reaches a given goal. Additionally, all these modules take in the set representation $\{P_1\ldots P_C\}$ as input to achieve compositional generalization.

\textbf{Set Feasibility Predictor} 
Similar to DiffSkill~\cite{lin2022diffskill}, we train a feasibility predictor $f_k(U^o, U^g)$ for each skill, where $U^o = \{u^o_i\}_{i=1\dots N_o}, U^g = \{u_j^g\}_{j=1\dots N_g}$ are latent set representations of an observation and goal point cloud respectively. The feasibility predictor outputs a value in $[0, 1]$ denoting if the goal can be reached from the observation by executing the $k^{th}$ skill. In DiffSkill, the feasibility predictor uses a flat representation that takes in a single latent vector for all objects in the scene as input. However, as our skills such as cutting or spreading only need to take in a subset of the objects as input, we use the same attention method for the feasibility and assume that the feasibility predictor only takes as input a subset of the full set representation $\hat{U}^o \subseteq U^o, \hat{U}^g \subseteq U^g$, where 
 $\hat{U}^o = \{u^o_i\}_{i=1\dots N_k},  \hat{U}^g = \{u_j^g\}_{j=1\dots M_k}$,
Here, $N_k$ and $M_k$ are the numbers of components in the observation and goal for skill $k$.  The number of components in the observation and goal  can be different since the number of components can change before and after executing a skill; for example, the cut skill takes one component as observation and cuts it into two components. As another example for robot assembly~\cite{lee2021ikea}, the number of entities increases when a piece is disassembled into parts and the number of entities decreases when the parts are assembled. In this work, we manually define the number of entities $N_k$, $M_k$ per skill. Determining which subset to attend to when executing each skill can be difficult; we make this decision during the planning and defer the details to Sec.~\ref{sec:planning}. We parameterize $f_k$ to be invariant to permutation using max-pooling layers. See Appendix for details.


We train the feasibility predictor of skill $k$ with positive examples $\hat{U}^o, \hat{U}^g$, where the goal $\hat{U}^g$ can be reached from the observation $\hat{U}^o$ within $T$ timesteps by executing skill $k$. The negative examples are goal $\hat{U}^g$ that cannot be reached from the observation $\hat{U}^o$ using skill $k$.
During training, we obtain positive pairs for the feasibility predictor by sampling two point clouds $(P^{obs}, P^{goal})$ from the same trajectory in the demonstration set. To find $\hat{U}^o, \hat{U}^g$, we first cluster the observation and goal point clouds into two sets $\{P^{o}_{i}\}, \{P^{g}_{j}\}$ respectively. Then, we match point clouds in the observation set to those in the goal set by finding the pairs of point clouds that are within a Chamfer distance of $\epsilon$: $\{(P^o_i, P^g_j) \mid D_{Chamfer}(P^o_i, P^g_j) < \epsilon \}$. We then remove these point clouds from the corresponding set, since these are the point clouds that have already been moved to the target location in the goal. We can then encode the remaining point clouds into $\hat{U}^o, \hat{U}^g$ as explained above. We generate hard negative samples by replacing one entity in the positive examples with a random latent vector.


\textbf{Set \rebuttal{Cost} Predictor}
As we do planning in a latent space, we train a set \rebuttal{cost} predictor as our planning objective which determines how close a plan is to a given goal. The set \rebuttal{cost} predictor $C$ takes two latent set representations as input $U^o, U^g$. Since our tasks focus on matching each entity in the observation with one in the goal, we assume they have the same number of components, i.e. $|U^o|= |U^g| = N_c.$ To compute the \rebuttal{cost}, we try to find the matching entity with the minimal matching cost:
$C\Big(\{u^{o}_{i}\}, \{u^{g}_{j}\}\Big) = \rebuttal{\argmin}_\sigma \sum_{i=1}^{N_c} c_\theta (u_i^o, u_{\sigma(i)}^g)$, 
where $\sigma$ is a permutation and $c_\theta$ is a \rebuttal{cost} prediction network parameterized by an MLP trained to predict the \rebuttal{Chamfer Distance between the point clouds corresponding to the two latent vectors. This allows faster planning compared to first decoding latent vectors to point clouds and then computing their distance.} \rebuttal{Finally, optimization of the cost is done by performing Hungarian matching between the two sets containing latent vectors.}
\subsection{Planning with Set Representation} \label{sec:planning}
Given an observation and a goal point cloud $P^{obs}, P^{goal}$, we plan for the types of skills to apply in sequence, the attention for each skill (i.e. find $\hat{U}^o \subseteq U^o$), and the latent subgoals for each skill (i.e. the exact value for each latent vector in $\hat{U}^o$). As the simplest approach, we run a three-level nested optimization: In the top level, we exhaustively search over the combinations of skills to apply at each step, i.e. $k_1\dots, k_H$, where $k_h$ indexes the skill applied at the high-level step $h$.  We only keep the sequences that end with the same set cardinality as the goal by ensuring that $\sum_{h=1}^H M_{k_h} - N_{k_h} = N_g - N_o,$ where $M_{k_h}$ and $N_{k_h}$ are the number of observation and goal components for the skill applied at step $h$ and $N_o$ and $N_g$ are the number of components in the observed and target point clouds. 

In the second-level optimization, we search over different attention structures. Denote the latent set at the high-level step $h$ to be $U^h$. We formally define the attention structure at step $h$ to be  $I^h$, which consists of a list of indices, each of length $N_{k_h}$, such that $I^h$ selects a subset from $U^{h-1}$ to be the input to the feasibility predictor, i.e.
$\hat{U}^{h-1} = U^{h-1}_{I^h} \subseteq U^{h-1}$. Assume that we have $N_h$ components before applying skill $k_h$, i.e. $|U^{h-1}|=N_h$ and skill $k_h$ takes $K_h$ components as its observation. We can search over all $C_{N_h}^{K_h}$ combinations of attention structures. For components not considered by the skill, its latent vector will remain the same at step $h$. The combination of each skill attention yields an attention structure \textbf{I} for the whole plan, as illustrated in Fig.~\ref{fig:overview}(c). 
For this level of optimization, we use a sampling-based procedure to avoid an exhaustive search over topologically equivalent attention structures. See the Appendix for how we do this efficiently.


In the low-level optimization, for each attention structure \textbf{I}, we follow the optimization in DiffSkill~\cite{lin2022diffskill}. We first sample multiple initializations for the set of latent subgoals \textbf{U}, where each latent vector in the set is initialized from our generative model. We can then perform gradient descent to further optimize the latent subgoals on the following objective:
\begin{equation}
    \operatorname*{arg\,min}_{\bf{k, I, U}} \rebuttal{J}(\mathbf{k}, \mathbf{I}, \mathbf{U}) = \prod_{h=1}^{H} f_{k_h}(\hat{U}^{h-1}, \hat{U}^h) \exp (\rebuttal{C(U^H, U^g)}),
    \label{eqn:latent_opt}
\end{equation}
where \textbf{k} is the skill sequence, \textbf{I} is the attention structure of the plan, \textbf{U} is the set of all latent subgoals, $\hat{U}^h = \{u_i^h\}_{i=1\dots M_{k_h}}$ are the latent subgoals at step $h$, $\hat{U}^0 = \hat{U}^o \subseteq U^o$ is the attended observed set, and $U^g$ is the goal set. 
Finally, we can use our policy to execute our plan by following each subgoal. A summary of our method can be found in Algorithm~\ref{algo:pasta}.

\SetKwComment{Comment}{/* }{ */}
\begin{algorithm}
\caption{Planning with Spatial and Temporal Abstraction (PASTA)}\label{alg:diff_plan}
\SetKwInOut{Input}{Input}
\SetKwInOut{Output}{output}
\Input{Demonstration Dataset $D_{demo}$, skill horizon $T$, planning horizon $H$, modules for neural skill abstraction $\pi_k, f_k, r_\theta$, Point Cloud VAE with encoder $\phi$, decoder $\psi$, prior $p_u$}
\label{algo:pasta}
\For {\text{each valid skill sequence} $k_1, \dots k_H$}{
    \For {\text{each valid attention structure} $I^1, \dots I^H$}{
    \text{Initialize different latent subgoals $\hat{U}^1, \dots \hat{U}^H$ from $p_u$} \;
    \text{Optimize}  latent subgoals $\hat{U}^1, \dots \hat{U}^H$  \text{according to Sec.~\ref{sec:planning}} 
    to \text{obtain cost} $\rebuttal{J}(\mathbf{k}, \mathbf{I}, \mathbf{U})$ \;
    }
}
\text{Choose skill sequence $\mathbf{k}$, attention structure $\mathbf{I}$, and subgoals $\mathbf{U}$ that minimizes $\rebuttal{J}(\mathbf{k}, \mathbf{I}, \mathbf{U})$} \;
\For{$h\gets0$ \KwTo $H-1$}{
    Decode the subgoal from $\hat{U}^h$ using the decoder $\psi$ \;
    Execute policy $\pi_{k_i}$ following the subgoal \;}
\end{algorithm}

\section{Experiments} 
\label{sec:exp_baseline}
Our experiments are categorized into three parts: In Sec.~\ref{sec:exp-setup}, we describe the experimental setups and the baselines we consider. In Sec.~\ref{sec:exp-baseline} and Sec.~\ref{sec:exp-ablation}, we show that PASTA outperforms the baselines and ablate different components in our framework. In Sec.~\ref{sec:exp-realworld}, we demonstrate that PASTA can be effectively transferred to the real world without any fine-tuning.

\subsection{Simulation Tasks and Baselines}
\label{sec:exp-setup}
\textbf{Environment setups} We consider several long-horizon dough manipulation tasks and divide them into two categories. First, we consider the three tasks from Diffskill~\cite{lin2022diffskill}: LiftSpread, GatherTransport, and CutRearrange. These tasks require the agent to sequentially compose at most two skills to spread, cut or transport the dough. We further propose two new generalization tasks: CutRearrangeSpread (CRS) and CRS-Twice, where there are more entities during testing than during training.
\rebuttal{Similar to prior work~\cite{lin2022diffskill}, we specify the minimal planning horizon required for each task. Our approach also succeeds when we increase the horizon up to twice as long. See the Appendix, Sec 4.5 for details.}

\textbf{Generalization tasks} The CRS task provides a number of demonstration trajectories performing one of the three skills: Cutting with a knife, pushing with a pusher, and spreading with a roller. The demonstration of each skill only shows a tool manipulating a single piece of dough. During testing, the agent needs to cut a dough into two, transport one piece to a spreading area, and then spread it. Generalization to more entities is required as there will be two entities in the scene, whereas during training there was only one entity. Can we do even more generalization? In the CRS-Twice task, we use the same agent trained on the CRS dataset and ask it to cut two pieces of dough from a chunk, transport both of them to a spreading area, and spread them both. This is a 6-horizon task with up to 3 entities in the scene, much more complex than the skill demonstrations the agent is trained on. Due to the long-horizon nature of CRS-Twice, we specify the skill skeleton and use receding horizon planning for all the planning-based methods. See the Appendix for details.

\textbf{Baselines}
We consider several baselines in simulation: First, a gradient-based trajectory optimizer with oracle information~(Traj-Opt), which can solve single-stage tasks for deformable object manipulation as shown in  prior works~\cite{huang2020plasticinelab}. Second, a model-free RL with Soft Actor Critic~\cite{haarnoja2018soft} with RGB-D image input~(SAC-Image). Third, a SAC agent takes in the dough, target dough, and tool point clouds as input, the same as our method (SAC-Point). Fourth is DiffSkill, a model-based planning method from Lin \emph{et al.}~\cite{lin2022diffskill}, which takes RGB-D images as input and has no spatial abstraction. The last one is Flat 3D, which extends DiffSkill to use 3D point clouds as input,  with a ``flat'' 3D representation that encodes the whole scene to a single latent vector without any spatial abstraction.  

\rebuttal{\textbf{Metric} We specify goals as 3D point clouds of different geometric shapes such as boxes and spheres at specific locations.} We report the normalized decrease in the Earth Mover Distance~(EMD) approximated by the Sinkhorn diverge~\cite{sejourne2019sinkhorn} computed as $s(t) = \frac{s_0 - s_t}{s_{0}},$ where $s_0, s_t$ are the initial and current EMD. We additionally set a threshold for the score to determine the success of a trial.

\subsection{Comparison with Baselines in Simulation}
\begin{table*}[ht]
    \captionsetup{font=small}
    \centering
    \scalebox{0.75}{

    \begin{tabular}{cccc|ccc}
    \toprule
     \multirow{2}{*}{\diagbox{Method}{Task (Horizon)}} & \multicolumn{3}{c|}{DiffSkill tasks} & \multicolumn{2}{c}{Generalization tasks} \\
      & LiftSpread (2) & GatherMove (2) & CutRearrange (3) & CRS (3) & CRS-Twice (6)  \\ 
    \midrule
    Traj-Opt (Oracle)~\cite{huang2020plasticinelab} & 0.818 / 40\% & 0.403 / 0\% & 0.511 / 20\% & 0.312 / 0\% & 0.227 / 0\%  \\
    SAC-Image~\cite{haarnoja2018soft} & 0.797 / 0\% & 0.567 / 20\% & 0.103 / 0\% & 0.562 / 0\% & 0.365 / 0\%  \\
    SAC-Point~\cite{haarnoja2018soft} & 0.796 / 0\% & 0.603 / 40\% & 0.147 / 0\% & 0.573 / 0\% & 0.353 / 0\%  \\
    DiffSkill-Image~\cite{lin2022diffskill} & \textbf{0.920} / \textbf{100\%} & 0.683 / 60\% & 0.249 / 20\% & -0.505 / 0\% & -  \\
    Flat 3D (Ours) & * & * &  0.797 / 60\% & -0.712 / 0\% & -0.108 / 0\% \\
    PASTA (Ours) & 0.904 / \textbf{100\%} & \textbf{0.715} / \textbf{100\%} &  \textbf{0.837} / \textbf{80\%} & \textbf{0.896} / \textbf{100\%} & \textbf{0.604 / 40\%}\\
    \bottomrule
    \end{tabular}}
    \caption{Normalized improvement and success rate of all methods on two sets of tasks: tasks in DiffSkill and tasks that require generalization to more steps and entities. For CRS and CRS-Twice, training data only contains skills operating on \textbf{one} component of dough but at test time there are more than two components. Only the best-performing baselines in CRS are evaluated on CRS-Twice. For LiftSpread and GatherMove, we consider the whole scene as a single spatial abstraction, so Flat 3D is equivalent to PASTA.}
    \label{tab:exp-main}
\end{table*}

\label{sec:exp-baseline}
Table~\ref{tab:exp-main} shows the quantitative results of simulation tasks. First, we show that using a 3D representation is beneficial to planning and complex manipulation, as PASTA matches DiffSkill in LiftSpread and outperforms it in all the other tasks. 
Second, we highlight \algo's compositional generalization power in CRS and CRS-Twice, in which there are additional components of dough at test time. Effectively using spatial abstraction to model the scene, PASTA achieves 100\% success rate in CRS and retains a good performance in CRS-Twice. All the baselines, especially the planning baselines (DiffSkill, Flat 3D) fail dramatically, as they can only produce plans that consist of scenes seen in training. Impressively, PASTA is the only approach that reaches a non-zero success rate on these tasks that require compositional generalization. \rebuttal{\algo~has a computational complexity of $\mathcal{O}(K^H\cdot|\bf{I}|\cdot T_G),$ where $K$ is the number of skills, $H$ is the planning horizon, $|\bf{I}|$ is the number of attention structures and $T_G$ is the time it takes to solve each gradient-based optimization. We discuss more efficient planning algorithms in the limitation section as well as in the Appendix, Sec. 5.}
\subsection{Ablation analysis}
\label{sec:exp-ablation}

\begin{wraptable}{r}{7cm}
    
    \captionsetup{font=small}
    \centering
    \scalebox{0.8}{
    \begin{tabular}{cc}
    \toprule
     Ablation Method & Performance / Success \\ 
    \midrule
    No Hard Negatives Feasibility  & 0.740 / 40\%  \\
    No Sampling Planning   &  -0.455 / 0\% \\
    No Gradient Planning   &  0.329 / 0\% \\
    PASTA (Ours)   & \textbf{0.837 / 80\%} \\
    \bottomrule
    \end{tabular}}

    \caption{Ablation results from CutRearrange.}
    \label{tab:ablation}
\end{wraptable}

Table~\ref{tab:ablation} shows the quantitative performance of each ablation and PASTA in CutRearrange.
First, we consider a variant of feasibility predictor's training, which removes the hard negative samples and only uses random negative sampling (No Hard Negatives), which halves the success rate. Second, we consider two variants of the planner, one without gradient-descent (No Gradient Planning) and one without sampling (No Sampling Planning). The results show that both components are crucial to planning. We provide more ablations on the policy and the planner in the Appendix. 
\begin{figure*}[t]
    \includegraphics[width=\textwidth]{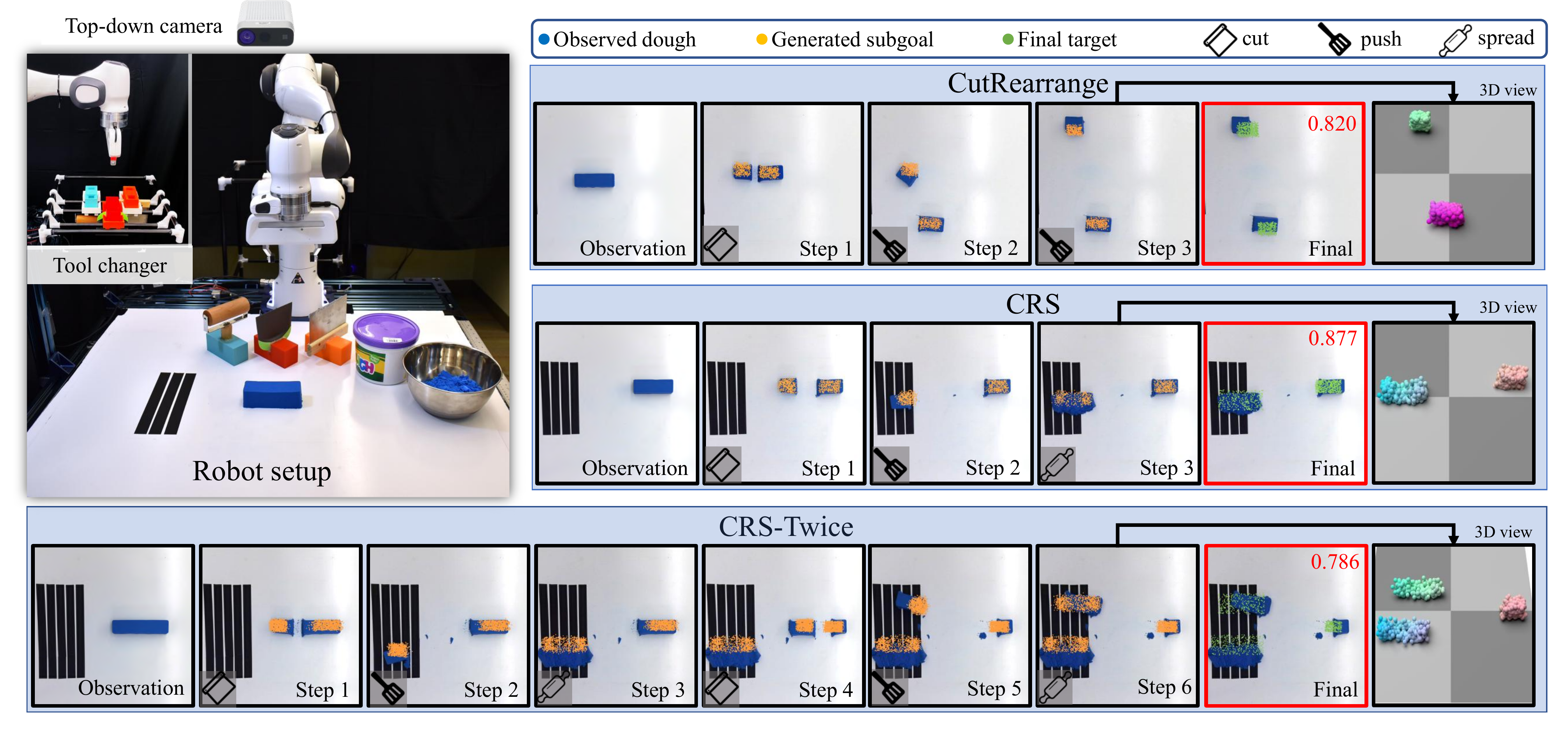}
     \caption{\textbf{Real world setup and execution with planned subgoals}. Our workspace consists of a Franka robot, a top-down camera, and a novel tool changer behind the robot that allows the robot to automatically switch tools. For each task, we show frames after executing a skill overlaid with the decoded point cloud subgoal; we report the final performance in red and overlay the ground truth target in green in the final frame. Additionally, we include a 3D view of the last generated subgoal to show the shape variations.}
     \label{fig:real-world}
\end{figure*}
 
\subsection{Real World Experiments}
\label{sec:exp-realworld}
Figure~\ref{fig:real-world} shows our real-world setup. We use a Franka robot with an Azure Kinect camera capturing the RGB-D observation of the workspace. The robot is equipped with a tool station~\cite{Zhang-2022-133215} that allows an automatic change of tools. For real-world ``dough'', we use Kinect Sand as a proxy because of its stable physical property. We transfer the feasibility predictor and \rebuttal{cost} predictor of PASTA directly from the simulation and define heuristic controllers for the skills. For evaluation, we first generate a desired target point cloud and then reset the dough to its initial shape and record its point cloud. We then plan using \algo~and then use the heuristic controllers to follow the plan. Finally, we report the normalized improvement EMD. We compare with the Flat3D method and also with human. 

We evaluate three of the simulation tasks: CutRearrange, CRS, and CRS-Twice. For each task, we evaluate the same four initial and target shapes for all methods and report the performance in Table~\ref{tab:realworld}. Figure~\ref{fig:real-world} shows the keyframes from the execution of PASTA. We overlay the planned subgoals as well as the final goal. PASTA performs on par with human in the real world, highlighting the robustness of our planner and the advantage of using 3D representation for sim2real transfer. 



\begin{table*}[ht]
    \captionsetup{font=small}
    \centering
    \scalebox{0.7}{
    \begin{tabular}{cccc}
    \toprule
     \diagbox{Method}{Task (Horizon)} & CutRearrange (3) & CRS (3) & CRS-Twice (6)  \\
    \midrule
    Flat 3D & 0.351 $\pm$ 0.478  & 0.007 $\pm$ 0.429 & - \\
    PASTA (Ours) & \textbf{0.836 $\pm$ 0.029}   & \textbf{0.854 $\pm$ 0.016} & \textbf{0.795 $\pm$ 0.035} \\
    \midrule
    Human &  0.910 $\pm$ 0.014 & 0.863 $\pm$ 0.018 & 0.895 $\pm$ 0.013 \\
    \bottomrule
    \end{tabular}}
    \caption{Normalized improvement on real-world tasks. Each entry shows the mean and std of the performance over 4 runs. Flat 3D does not produce any meaningful plan for CRS, so we do not evaluate it on CRS-Twice.}
    \label{tab:realworld}
\end{table*}
\section{Conclusions and Limitations}
In this work, we propose a planning framework named PASTA that incorporates both spatial and temporal abstraction by planning with a 3D latent set representation with attention structure. We demonstrate a manipulation system in the real world that uses PASTA to plan with multiple tool-use skills to solve the challenging deformable object manipulation tasks, and we show that it significantly outperforms a flat 3D representation, especially when generalizing to more complex tasks.

%
\rebuttal{\textbf{Limitations:} First, we only transfer the planner to the real world and use heuristic controllers instead of the policy trained in simulation. This is due to the sim2real gap caused by the differences in dough's physical parameters, table friction, and occlusions from the robot arm. Prior work~\cite{qi2022ral} shows promising results in transferring a closed-loop policy taking partial point clouds as input, and future work can explore better sim2real methods. Second, our planner exhaustively searches overall skill combinations and attention structures and does not scale well to longer sequences with more skills. Future work can incorporate more efficient search algorithms or priors to prune the search space. Finally, we rely on an unsupervised clustering method for entity decomposition and a point cloud VAE for mapping an observation to our latent set representation. Future work can incorporate self-supervised methods for learning the decomposition. For further discussion on the limitations and future work, please see Sec. 5 of the Appendix.}

\label{sec:conclusion}




\acknowledgments{This material is based upon work supported by the National Science Foundation under Grant No. IIS-2046491, IIS-1849154, NSF award under AWD00001520, LG Electronics, Carnegie Mellon University’s GSA/Provost GuSH Grant funding, and Amazon Research Award.}


\bibliography{ref}  

\end{document}


\begin{center}
{\huge \textbf{Appendix}}
\end{center}

\tableofcontents
\newpage

\setcounter{section}{0}
\section{Implementation Details}
\subsection{Entity Encoding}
To train our point cloud variational autoencoder~\cite{pointflow}, we normalize the point cloud of each entity $P_i$ to be centered at the origin, i.e. $\bar{P}_i = P_i - t_i$, where $t_i$ is the mean of all points in $P_i$. We then encode each centered $\bar{P}_i$ into a latent encoding: $z_i = \phi(\bar{P}_i)$. Our latent representation $u_i = [z_i, t_i]$ consists of the encoding of the point cloud's shape $z_i$ and the position of the center of the point cloud $t_i$. We model the point cloud's position $t_i$ explicitly such that the learned latent embedding $z_i$ can focus on shape variation alone and the model that plans over $u_i$ can still reason over point clouds at different spatial locations. During training, we record the 3D bounding box of all training data $t_{min}, t_{max} \in \mathbb{R}^3$, and we sample from $[t_{min}, t_{max}]$ during planning. We denote this combined distribution of $u=[z,t]$ as $P_u$.

\subsection{Details on Training Set Feasibility Predictor}
\label{sec:detail-fea}
\textbf{Hard Negative Samples} Suppose skill $k$ takes $N_k$ latent vectors from observation $\hat U^o$ and $M_k$ latent vectors from goal $\hat U^g$ as input. To generate random pairs of observations and goals as negative samples for the feasibility predictor, we can sample each latent point cloud representation $u_i$ by sampling the shape $z_i$ from the VAE prior $p_z$ and sampling the position $t_i$ from the distribution of positions in the training dataset. Such random negative samples are used similarly in DiffSkill~\cite{lin2022diffskill}. However, as the combined dimension of the set representation becomes larger compared to a flat representation, we need a way to generate harder negative samples. To do so, for a positive pair of set representation $(\{u^{o}_{i}\}, \{u^{g}_{j}\})$, we randomly replace one of the entities $u^o_i$ or $u^g_j$ with a random sample in the latent space and use it as a negative sample. Our ablation results show that this way of generating hard negative samples is crucial for training our set feasibility predictor.

\textbf{Noise on Latent Vectors} During the training of the feasibility predictor, for each of the input latent vector $u = [z, t]$, where $z \in \mathbb{R}^{D_z}$ is the latent encoding of the shape and $t \in R^3$ is the 3D position of the point cloud, we add a Gaussian noise to each part, i.e. $\hat{z} = z + \sigma_z \epsilon$ and $\hat{t} = t + \sigma_t \epsilon$, where $\epsilon \sim \mathcal{N}(\mathbf{0}, \mathbf{I})$.  The amount of noise determines the smoothness of the feasibility landscape. Without any noise, planning with gradient descent with the feasibility function becomes much harder.

\subsection{Details on the Attention Structure for Planning with Set Representation}
Given the initial latent set observation $U^{obs}$ with $N_o$ components, and the skill sequences $k_1, \dots, k_H$, in this section we describe how to generate the attention structure. We denote the latent set representation at step $h$ as $U^h$, $h=1 \dots H$ and define $U^0 = U^{obs}$. As skill $k_h$ takes in $N_{k_h}$ components as input and $M_{k_h}$ components as output, by calculation we know that $U^h$ has $N_o + \sum_{i=1}^h M_{k_i} - N_{k_i}$ components. From now on, we denote $|U^h|=N_h$ and $U^h = \{u_{h,1}, \dots, u_{h,N_h}\}$. As the skill $k_h$ only applies to a subset of the input $U^{h-1}$, we now formally define the attention structure at step $h$ to be  $I^h$, which consists of a list of indices, each of length $N_{k_h}$, such that $I^h$ selects a subset from $U^{h-1}$ to be the input to the feasibility predictor, i.e.
\begin{equation*}
    \hat{U}^{h-1} = U^{h-1}_{I^h} \subseteq U^{h-1}.
\end{equation*}

However, enumerating all $I^h$ is infeasible, as there are $C_{N_h}^{N_{k_h}}$ combinations for each step. Fortunately, we do not have to enumerate all different structures. The insight here is that for each attention structure $I_1, \dots I_H$, we will perform a low-level optimization. In this low-level optimization, we will first initialize all the latent vectors to be optimized from $P_u$ and then perform gradient descent on them. As many of the attention structures yield topologically equivalent tree structures (An example of such a tree is illustrated in Figure 2c of the main paper), and each latent vector in the tree is sampled independently from the same distribution $P_u$, these topologically equivalent tree structures result in the same optimization process. As such, we do not need to exhaust all of such attention structures.

Instead of enumerating each topologically different structure and then sampling multiple initializations for the low-level optimization, we randomly sample sequences $(I_1, \dots I_H)$ and perform low-level gradient-descent optimization on all the samples. In this way, with enough samples, we will be able to cover all attention structures.

Now, we can sequentially build up the subgoals latent set representation during planning. Specifically, assuming that we have constructed the previous latent set representation $U^{h-1}$, we will now describe the procedure for constructing $U^h,$ as well as the predicted feasibility for the current skill $k_h$, i.e. $f_{k_h}(\hat{U}^{o, h}, \hat{U}^{g, h}),$ where $\hat{U}^{o, h}, \hat{U}^{g, h}$ are the subset of $U^{h-1}$ and $U^{h}$ attended by the feasibility predictor. First, we generate $I_h$ by randomly choosing the index of a subset from $U_{h-1}$. $U^h$ are composed of two parts: The first part is the latent vectors generated by applying the skill $k_h$. For this part, we will create a set of new vectors $u_{h, 0}, \dots u_{h, M_{h_k}}.$ This part of the latent vectors will be attended by the feasibility predictor as $\hat{U}^{g, h} = \{u_{h, 0}, \dots u_{h, M_{h_k}}\}$. The second part of $U^h$ comes from the previous latent set vectors that are not modified by the skill, i.e. $U^{h-1} \setminus U^{h-1}_{I_h}$, and $U^h$ is the addition of both parts, i.e.
\begin{equation*}
    U^h = \hat{U}^{g, h} \cup (U^{h-1} \setminus U^{h-1}_{I_h})
\end{equation*}

In this way, we can sequentially build up $U^h$ from $U^{h-1}$, and $U^{0}$ is simply $U^{obs}$. At the same time, we have determined our attention structure and the feasibility prediction. Our objective can thus be written as
\begin{equation}
    \operatorname*{arg\,min}_{\bf{k}, \bf{I}, \bf{U}} J(\mathbf{k}, \mathbf{I}, \mathbf{U}) = \prod_{h=1}^{H} f_{k_h}(\hat{U}^{o, h}, \hat{U}^{g, h}) \exp (-C(U^H, U^g)),
    \label{eqn:latent_opt_app}
\end{equation}
where $\bf{U}$ is the set union of all latent vectors to be optimized.

\subsection{Network Architectures}
\textbf{Set Feasibility Predictor} We use a Multi-Layer Perceptron (MLP) with ReLU activations for our feasibility predictor. We apply max-pooling to the transformed latent vectors of $\hat U^o$ and $\hat U^g$ to achieve permutation invariance. Below is our architecture:
\begin{figure}[ht]
    \centering
    \includegraphics[width=0.6\textwidth]{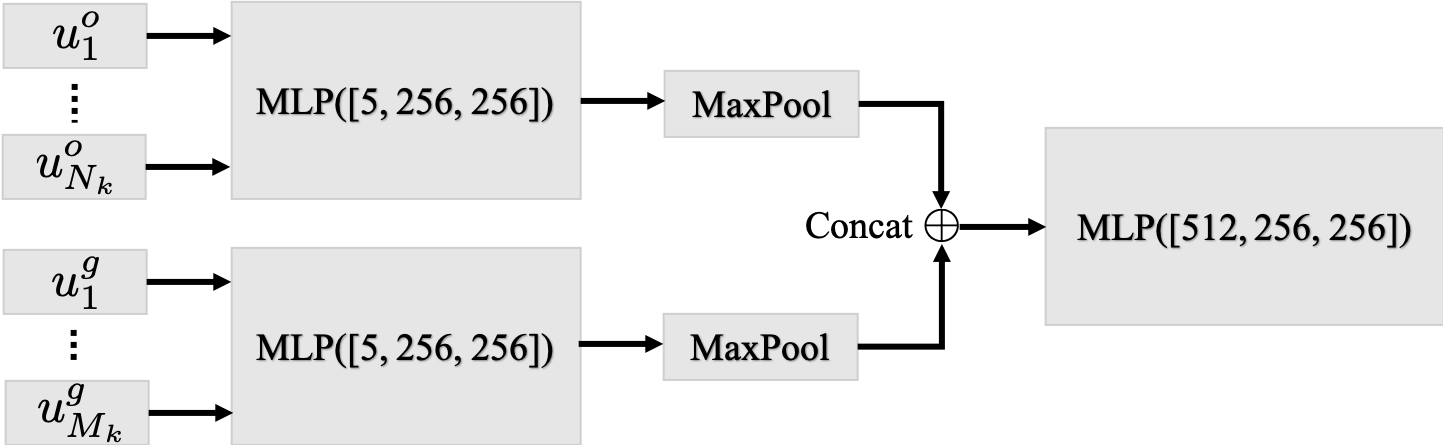}
    \caption{Architecture for the set feasibility predictor}
    \label{fig:fea-arch}
\end{figure}

\textbf{Set Cost Predictor} We use a 3-layer MLP with a hidden dimension of $1024$ and ReLU activations.

\textbf{Set Policy}
The set point cloud policy for the $k^{th}$ skill $\pi_k$ takes in an observed point cloud $P^{obs}$, a goal point cloud $P^{goal}$, and a tool point cloud $P^{tool}_{k}$ and outputs an action at each timestep to control the tool directly. The tool point cloud $P^{tool}$ is obtained by sampling points on the mesh surface of the tool and then transforming these points to the same camera frame as the $P^{obs}$ and $P^{goal}$, assuming the pose of the tool is known from the robot state. Instead of taking latent vectors as input, the policy functions directly in point cloud space, which allows it to handle times when spatial abstraction is ambiguous. For instance, during cutting and merging, the number of dough components gradually increases or decreases, during which the latent set representation is not changing smoothly while the point clouds change smoothly during the process. We later show the advantage of using point clouds directly as the policy input. We concatenate each point's $(x,y,z)$ coordinates with a  one-hot encoding to indicate whether the point belongs to the observation, tool, or goal, and we input the points into a PointNet++~\cite{qi2017pointnet++} encoder followed by an MLP which outputs the action. We use a point cloud for the tool to allow the PointNet encoder to reason about the interaction between the tool and the dough in the same space.
We use PyTorch Geometric's \cite{Fey/Lenssen/2019} implementation of PointNet++ and with the following list of modules in our encoder.
\begin{verbatim}
SAModule(0.5,0.05,MLP([3+3,64,64,128]))
SAModule(0.25,0.1,MLP([128+3,128,128,256]))
GlobalSAModule(MLP([256+3,256,128,512,1024]))
\end{verbatim}
The MLP following the encoder consists of hidden dimensions $[1024, 512, 256]$ and ReLU activations.

\rebuttal{
\begin{table}[t]
    \centering
    \begin{tabular}{lcccc}
    \toprule
      &  LiftSpread & GatherMove & CutRearrange & CRS + CRS-Twice  \\
    \midrule
    \# of initial configurations & 200 & 200 & 1500 & 1200 \\
    \# of target configurations & 200 & 200 & 1500 & 1200 \\
    \# of training trajectories & 1800 & 1800 & 1350 & 1080 \\
    \# of testing trajectories  & 200 & 200 & 150 & 120 \\
    \# of total trajectories & 2000 & 2000 & 1500 & 1200 \\
    \# of total transitions & 1e5 & 1e5 & 7.5e4 & 6e4 \\
    \bottomrule
    \end{tabular}
    \vspace{2mm}
    \caption{Summary of training/testing data}
    \label{tab:train-data}
\end{table}

\subsection{Training Details}

\textbf{Training data.} We inherit the data generation procedure from DiffSkill~\cite{lin2022diffskill}: first, we randomly generate initial and target configurations. The variations in these configurations include the location, shape, and size of the dough and the location of the tool. We then sample a specific initial configuration and a target configuration and perform gradient-based trajectory optimization to obtain demonstration data. For each task, the demonstration data consists of all the transitions from executing the actions outputted by the trajectory optimizer. We perform a train/test split on the dataset and select 5 configurations in the test split for evaluating the performance for all the methods. More information about training and testing data can be found in Table~\ref{tab:train-data}.

\textbf{Point cloud VAE.} We train our point cloud VAE by maximizing the evidence lower bound (ELBO). For a dataset of observations $P(X)$, which consists of the segmented point cloud of each entity in the scene, we optimize the following objective:
\begin{equation}
    \mathcal{L}_{VAE} = \mathbb{E}_{Q_{\phi}(z|x)} \left[
    \log P_{\psi}(X | z)
    \right] - D_{KL}(Q_{\phi}(z|X) || p(z))
\end{equation}
where $Q_{\phi}(z|X)$ is the encoder modeled as a diagonal Gaussian, $P_{\psi}(X|z)$ is the decoder, and $p(z)$ is standard Gaussian. The VAE is pre-trained, and we fix its weights prior to training the other modules.

\textbf{Point cloud policy.} We train our point cloud policy with standard behavioral cloning (BC) loss, i.e. for the $k$-th skill, we sample a transition from the demonstration data, which contains the observed point clouds $\{P^{o}_i\}$, goal point clouds $\{P^{g}_{j}\}$, a tool point cloud $P^{tool}_{k}$, and the action of the tool $a$. Then, we match point clouds in the observation set to those in the goal set by finding the pairs of point clouds that are within a Chamfer Distance of $\epsilon$: $\{(P^o_i, P^g_j) \mid D_{Chamfer}(P^o_i, P^g_j) < \epsilon \}$ and filter out the non-relevant point clouds. Last, we pass the filtered point clouds into the policy and minimize the following loss:
\begin{equation}
    \mathcal{L}_{\pi_k} = \mathbb{E} \left[ \|a - \pi_{k}(\{P^o_i\}, P^{tool}_{k}, \{P^g_j\})\|^2 \right]
\end{equation}

\textbf{Feasibility predictor.} We train the feasibility predictor for the $k$-th skill by regressing to the ground-truth feasibility label using mean squared error (MSE) as loss, i.e.
\begin{equation}
    \mathcal{L}_{f_k} = \mathbb{E}\left[
    \left(f_k(\hat{U}^o, \hat{U}^g) - \mathbbm{1} \{\hat{U}^o, \hat{U}^g~\text{is a positive pair}\}
    \right)^2\right]
\end{equation}
During training, we obtain positive pairs for the feasibility predictor by sampling two point clouds $(P^{obs}, P^{goal})$ from the same trajectory in the demonstration set. To find $\hat{U}^o, \hat{U}^g$, we first cluster the observation and goal point clouds into two sets $\{P^{o}_{i}\}, \{P^{g}_{j}\}$ respectively. Then, we match point clouds in the observation set to those in the goal set by finding the pairs of point clouds that are within a Chamfer Distance of $\epsilon$: $\{(P^o_i, P^g_j) \mid D_{Chamfer}(P^o_i, P^g_j) < \epsilon \}$. We then remove these point clouds from the corresponding set, since these are the point clouds that have already been moved to the target location in the goal. We can then encode the remaining point clouds into $\hat{U}^o, \hat{U}^g$ using our VAE.

\textbf{Cost predictor.} We train the cost predictor by simply regressing to the Chamfer Distance (CD) between two entities represented by their latent vectors, i.e.
\begin{equation}
    \mathcal{L}_c = \mathbb{E}\left[
    \left(c(\phi(P_i), \phi(P_j)) - D_{Chamfer}(P_i, P_j)\right)^2\right]
\end{equation}
where $P_i$ and $P_j$ are point clouds of a single entity sampled from the dataset and $\phi$ is the encoder. There are two reasons that we train a cost predictor on latent vectors instead of directly computing the Chamfer Distance between two point clouds. For one, decoding each latent vector would greatly bottleneck the planning speed. Experiments on CutRearrange show that with our learned cost predictor, the planning takes 35s; on the other hand, if we decode the latent vectors and use the Chamfer Distance, even with a subsampled point cloud of 200 points, the planning takes 37200s (around 10 hours), which is impractical to use. Moreover, using a cost predictor can also offer us the flexibility to incorporate more complex reward functions in the future.

Finally, We train our policy, feasibility predictor, and cost predictor  with the following loss:
\begin{equation}
    \mathcal{L}_{PASTA} = \sum_{k=1}^K \mathbb{E} \left[
     \lambda_\pi \mathcal{L}_{\pi_k} +
     \lambda_f \mathcal{L}_{f_k} +
     \lambda_c \mathcal{L}_c
    \right]
\end{equation}
We use $\lambda_\pi=1$, $\lambda_f = 10$, and $\lambda_c = 1$ for all of our experiments.
}

\section{Details on Simulation Experiments}

\subsection{Hyperparameters for Simulation Dough}
We use PlasticineLab~\cite{huang2020plasticinelab} for evaluating our simulation experiments. We provide the hyperparameters that are relevant to the properties of the dough in simulation to enhance the replicability of our results. See Table~\ref{tab:sim-dough} for details.

\begin{table}[t]
    \centering
    \begin{tabular}{ccccc}
    \toprule
    Parameter  &  LiftSpread & GatherMove & CutRearrange & CRS + CRS-Twice  \\
    \midrule
    Yield stress & 200 & 200 & 150 & 150 \\
    Ground friction & 1.5 & 1.5 & 0.5 & 0.5 \\
    Young's modulus (E)   & 5e3 & 5e3 & 5e3 & 5e3\\
    Poisson's ratio ($\nu$) & 0.15 & 0.15 & 0.15 & 0.15\\
    \bottomrule
    \end{tabular}
    \vspace{2mm}
    \caption{Parameters for simulation dough}
    \label{tab:sim-dough}
\end{table}

\subsection{Hyperparameters for DBSCAN}
To cluster a point cloud, we use Scikit-learn's~\cite{scikit-learn} implementation of DBSCAN~\cite{ester1996density} with \verb|eps=0.03, min_samples=6, min_points=10| for all of our environments. Further, we assign each noise point identified by DBSCAN to its closest cluster.

\subsection{Hyperparameters for PASTA}
Table~\ref{tab:pasta-hyper} shows the hyperparameters used for PASTA in our simulation tasks. Planning for CRS-Twice requires a large number of samples. Therefore, we modify the planner to improve sample efficiency. See Sec.~\ref{sec:plan-crs2} for details.

\begin{table*}[ht]
\centering
\begin{tabular}{@{}lccccc}
\toprule
Training parameters & LiftSpread & GatherMove & CutRearrange & CRS & CRS-Twice  \\
\midrule
\textit{Point Cloud VAE} & & & &\\
\hspace{5mm}learning rate & 2e-3 & 2e-3 & 2e-3 & 2e-3 & 2e-3\\
\hspace{5mm}latent dimension & 2 & 2 & 2 & 2 & 2\\
\hspace{5mm}number of parallel GPUs & 4 & 4 & 8 & 10 & 10\\
\hspace{5mm}number of training epochs & 8 & 4 & 100 & 11 & 11\\
\textit{Feasibility predictor} & & & &\\
\hspace{5mm}learning rate & 1e-4 & 1e-3 & 1e-4 & 1e-4 & 1e-4\\
\hspace{5mm}batch size & 256 & 256 & 256 & 256 & 256 \\
\hspace{5mm}noise on shape encoding $\sigma_z$ & 0 & 0 & 0 & 0.02 & 0.02 \\
\hspace{5mm}noise on position $\sigma_t$ & 0.01 & 0.01 & 0.005 & 0.01 & 0.01 \\
\textit{Cost predictor} & & & &\\
\hspace{5mm}learning rate & 1e-4 & 1e-3 & 1e-4 & 1e-4 & 1e-4\\
\hspace{5mm}batch size & 256 & 256 & 256 & 256 & 256 \\
\textit{Policy} & & & &\\
\hspace{5mm}learning rate & 1e-4 & 1e-3 & 1e-4 & 1e-4 & 1e-4\\
\hspace{5mm}batch size & 10 & 10 & 10 & 10 & 10 \\
\hspace{5mm}noise on point cloud & 0.005 & 0.005 & 0.005 & 0.005 & 0.005 \\
\toprule
Planning parameters &  LiftSpread & GatherMove & CutRearrange & CRS & CRS-Twice  \\
\midrule
    \hspace{5mm}learning rate & 0.01 & 0.01 & 0.01 & 0.01 & 0.01\\
    \hspace{5mm}number of iterations & 200 & 100 & 100 & 200 & 300 \\
    \hspace{5mm}number of samples & 5000 & 5000 & 5000 & 50000 & 500000\\
\bottomrule
\end{tabular}
\caption{Summary of hyperparameters used in PASTA. For CRS-Twice, we use the same model as CRS but modify the planner to have better sample efficiency.}
\label{tab:pasta-hyper}
\end{table*}

\subsection{Receding Horizon Planning for CRS-Twice}
\label{sec:plan-crs2}
As the planning horizon increases, the number of possible skill sequences as well as the number of possible attention structures increases exponentially. The task of CRS-Twice has a planning horizon of 6 and is a much more difficult task to solve. As such, for this task, we specify the skill sequences and use Receding Horizon Planning (RHP). Starting from the first time step, we follow the procedure in Algorithm 1 but only optimize for $H_{RHP}$ steps into the future and compare the achieved subgoal with the final target to compute the planning loss. After optimization, we take the first subgoal from the plan and discard the rest of the plan. We then repeat this process until we reach the overall planning horizon $H$. In our experiments, we use $H_{RHP}=3$. While we can perform model predictive control and execute the first step before planning for the second step, we find this open-loop planning and execution to be sufficient for the task.

\section{Details on Real World Experiments}
\subsection{Heuristic Policies}
Transferring the learned policies from simulation to the real world can be more difficult than transferring the planner itself, as the policies are affected more by the sim2real gap, such as the difference in friction and properties of dough in the real world. To sidestep this challenge, for our real world experiments, we design three heuristic policies: cut, push, and roll to execute the generated plans in the real world.

Just like our learned policies in simulation, each heuristic policy takes in the current observation and the generated subgoal in point clouds and outputs a sequence of desired end effector positions used for impedance control. In addition, each policy takes in the attention mask provided by the planner indicating the components of interest. The same DBSCAN procedure is used for this. The cut policy first calculates the cutting point by computing the length ratio of the resulting components. Then it cuts the dough and separates it such that the center of mass of each resulting component matches the one in the subgoal. For the push policy, given a component and a goal component, the policy pushes the dough in the direction that connects the two components' center of mass. The roll policy first moves the roller down to make contact with the dough. Then, based on the goal component's length, the policy calculates the distance it needs to move the roller back and forth when making contact with the dough.

\subsection{Procedure for Resetting the Dough}\label{reset}
To compare different methods with the same initial and target configurations, we  first use a 3D-printed mold to fit the dough to the same initial shape. We then overlay the desired initial location on the image captured by the top-down camera and place the dough at the corresponding location in the workspace to ensure different methods start from the same initial location.

\subsection{Procedure for the Human Baseline}
Following the same procedure in section \ref{reset}, we first reset the dough to the initial configuration. Then, we overlay the goal point cloud on the image captured by the top-down camera. The overlay image is shown on a screen and presented to the human in real time when the human is completing the task.

\subsection{Procedure for Making the Real Dough}
\begin{table}[h!]
    \centering
    \begin{tabular}{ccc}
    \toprule
    Material     &  Quantity(g) & Baker's percentage(\%)   \\
    \midrule
    Flour    & 300 & 100 \\
    Water  & 180 & 60\\
    Yeast & 3 & 1 \\
    \bottomrule
    \end{tabular}
    \vspace{2mm}
    \caption{All-purpose dough recipe}
    \label{tab:recipe}
\end{table}
We follow the recipe shown in Table~\ref{tab:recipe} to make the real dough. Following the tradition of baking, we use the backer's percentage, so that each ingredient in a formula is expressed as a percentage of the flour weight, and the flour weight is always expressed as 100\%. First, we take 300 grams of flour, 3 grams of yeast, and 180 grams of water into a basin. Then, we mix the ingredients and knead the dough for a few minutes. Next, we use a food warp to seal the dough in the basin and put them in the refrigerator to let the dough rest for 4-5 hours. Finally, we take out the dough from the refrigerator and reheat it with a microwave for 30-60 seconds to soften it.

\section{Additional Experiments}
\subsection{Ablation Studies}
\begin{wraptable}{r}{7cm}
    \captionsetup{font=small}
    \centering
    \scalebox{0.9}{
    \begin{tabular}{cc}
    \toprule
     Ablation Method & Performance / Success \\
    \midrule
    No Smoothing Feasibility  & 0.744 / 40\%  \\
    Shared Encoder Policy & -0.304 / 0\%  \\
    Tool Concat Policy & 0.516 / 60\%  \\
    Set without Filtering & 0.360 / 20\%  \\
    PASTA (Ours)   & \textbf{0.837 / 80\%} \\
    \bottomrule
    \end{tabular}}
    \caption{Additional ablation results from CutRearrange.}
    \label{tab:app-ablation}
\end{wraptable}
\textbf{Ablations on feasibility predictor.} Following the discussions in Sec~\ref{sec:detail-fea}, we train a feasibility predictor without adding any noise to show that adding noise helps with the optimization landscape during planning. We call this ablation \emph{No Smoothing Feasibility}. As shown in Table~\ref{tab:app-ablation}, this variant only achieves half of the success rate of PASTA, suggesting the importance of noise during training.

\textbf{Ablations on policy.} We consider two ablation methods for our set policy. First, we consider a \emph{Shared Encoder Policy} that takes in the latent vectors from the encoder and uses a max pooling layer followed by an MLP to produce the action. The architecture is very similar to our Set Feasibility Predictor. Our results in Table~\ref{tab:app-ablation} show that this architecture has zero success in our task. We hypothesize that this is because the entity encoding can be unstable during the skill execution. For example, during cutting, the dough slowly transitions from one piece to two pieces, making the input to the policy unstable.

Second, we compare with a \emph{Tool Concat Policy} that takes in the observation and goal point cloud of the dough, passes them through a PointNet++~\cite{qi2017pointnet++} encoder to produce a feature, and then concatenates the tool state to the feature. The concatenated feature is passed through a final MLP to output the action. In comparison, the set policy in PASTA takes the point cloud of the tool and concatenates it with the dough in the point cloud space before passing it to the PointNet. We hypothesize that this way allows PointNet to reason more easily about the spatial relationships between the tool point cloud and the dough point cloud. Results in Table~\ref{tab:app-ablation} highlight the advantage of using a point cloud to represent the tool.

\textbf{Ablation on set representation.} We consider a variant of PASTA \emph{Set without Filtering}, which uses the same set representation as PASTA, but does not filter entities that are approximately the same both during training and testing. This filtering is only possible with a set representation and we want to show the advantage of this filtering. For this ablation, during training, the feasibility predictor takes in all the entities in the scene in set representation, and the policy takes in the concatenation of point clouds from each entity. During planning, we do not enumerate attention structures but instead optimize for all the entities. As shown in Table~\ref{tab:app-ablation}, without filtering, this ablation performs significantly worse than PASTA, showing that filtering is an important advantage enabled by our set representation.

\subsection{Visualization of the Latent Space}

We visualize the latent space of PASTA in CutRearrange in Figure~\ref{fig:latent1} and visualize the latent space of Flat 3D baseline in Figure~\ref{fig:latent5} for comparison. Since we use a latent dimension of 2 for all of our environments, we can visualize the original latent space without applying any dimensionality reduction techniques. PASTA only encodes the shape of each entity and thus can better model the variations in shapes. On the other hand, Flat 3D couples the shape variation with the relative position of two entities. This makes a flat representation difficult to generalize compositionally to scenes with different numbers of entities or scenes with entities that have novel relative spatial locations to each other.

\begin{figure}[t]
     \centering
     \begin{subfigure}[b]{0.45\textwidth}
         \centering
         \includegraphics[width=\textwidth]{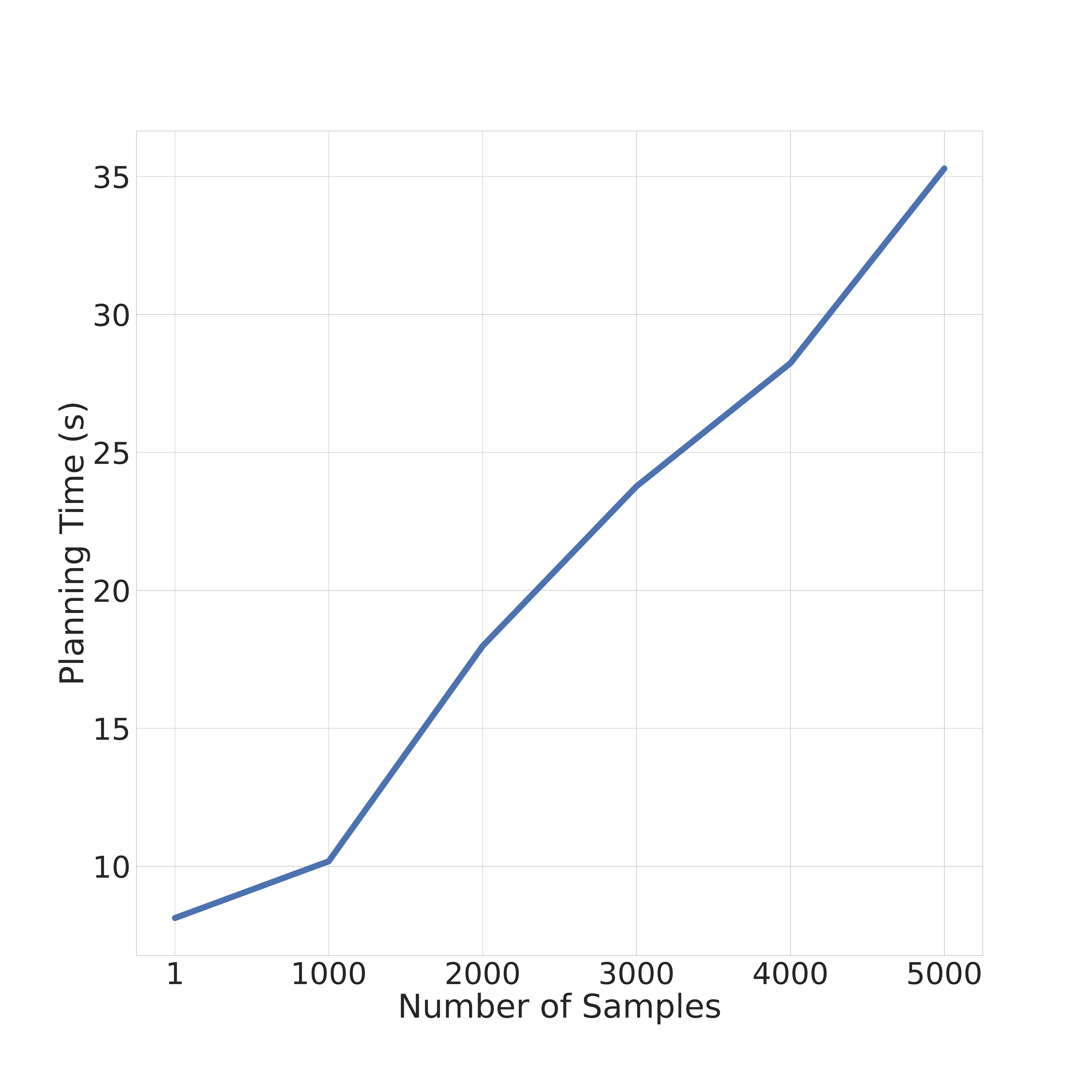}
         \caption{Planning time v.s. number of samples}
         \label{fig:time1}
     \end{subfigure}
     \hfill
     \begin{subfigure}[b]{0.45\textwidth}
         \centering
         \includegraphics[width=\textwidth]{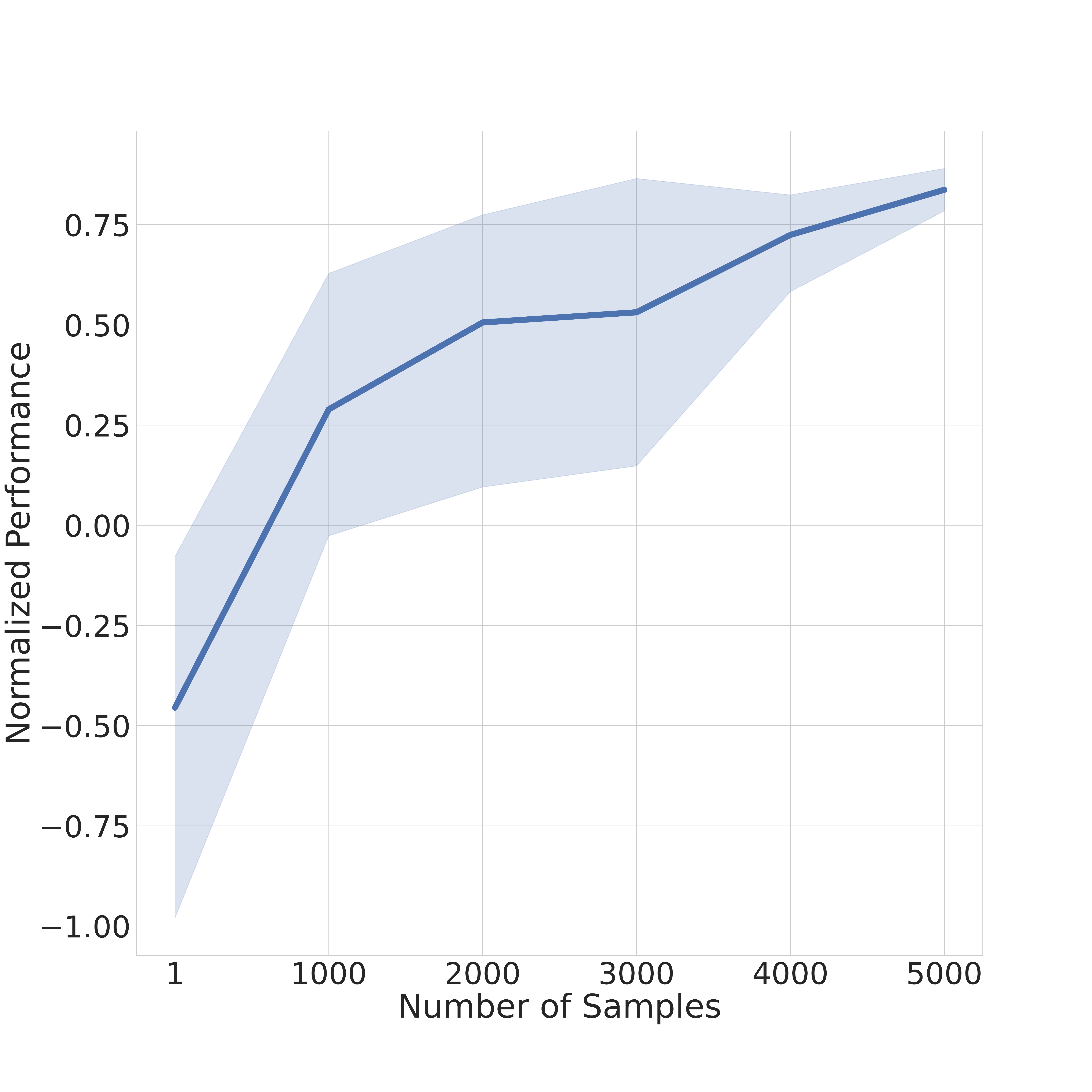}
         \caption{Planning performance v.s. number of samples}
         \label{fig:time2}
     \end{subfigure}
        \caption{Planning time and performance with varying number of samples in CutRearrange.We show the mean and standard deviation of performance over 5 runs.}
        \label{fig:time}
\end{figure}

\subsection{Runtime of PASTA}
We implement the planning in the latent set representation in an efficient way, which can plan with multiple different structures in parallel on a GPU. To demonstrate the efficiency of PASTA, we vary the number of samples used for planning and record the planning time and final performance. We conduct the experiments in CutRearrange. Figure~\ref{fig:time1} shows that the planning time scales approximately linearly with the number of samples, and Figure~\ref{fig:time2} shows the planning performance versus the number of samples. As the result suggests, PASTA can achieve its optimal performance with a very short amount of planning time (under 1 minute) for the majority of our tasks. Finally, we summarize the planning time for all of our tasks in simulation in Table~\ref{tab:pasta-time}.
\begin{table*}[ht]
\centering
\begin{tabular}{@{}lccccc}
\toprule
 & LiftSpread & GatherMove & CutRearrange & CRS & CRS-Twice*  \\
\midrule
    \hspace{5mm}Planning time (seconds) & 58 & 35 & 35 & 307 & 7810\\
\bottomrule
\end{tabular}
\caption{Summary of planning time of PASTA in all of the simulation tasks. CRS-Twice uses Receding Horizon Planning, which results in an increase in planning time.}
\label{tab:pasta-time}
\end{table*}

\rebuttal{
\subsection{Additional metrics for real-world experiments}
We also quantitatively computed the action error v.s. subgoal error for our real-world trajectories. The results are shown in Table~\ref{tab:action-error}. From the results in the table, our planned goal is closer to the ground truth goal than the achieved goal, measured by the Earth Mover’s Distance (EMD), which shows that the controller does not compensate for the error of the planner.

\begin{table*}[ht]
\centering
\begin{tabular}{@{}lccc}
\toprule
 & CutRearrange & CRS & CRS-Twice  \\
\midrule
    \hspace{5mm} EMD(planned goal, ground-truth goal) & 0.038 $\pm$ 0.004 & 0.027 $\pm$ 0.004 & 0.029 $\pm$ 0.002 \\
    \hspace{5mm} EMD(reached goal, ground-truth goal) & 0.056 $\pm$ 0.007 & 0.044 $\pm$ 0.006 & 0.054 $\pm$ 0.016 \\
\bottomrule
\end{tabular}
\caption{Action error v.s. subgoal error for real world experiments. For each task, the mean $\pm$ std for 4 trajectories are shown.}
\label{tab:action-error}
\end{table*}

\subsection{Robustness of PASTA}
We show that \algo~is robust to two types of variations and can retain high performance.

\textbf{Robust to planning horizon} First, we increase the planning horizon from the minimal length for the task (3) to twice the minimal length (6), and we observe that \algo~retains a high performance across all horizons. The results are shown in Table~\ref{tab:vary-horizon}. This suggests that in practice, one can specify a maximum planning horizon for \algo~when the exact horizon is unknown.

\textbf{Robust to distractors} Second, we show that \algo~is robust to distractors in the scene. We add 2 distractor objects in CRS (which makes the scene have 4 objects in total). We observe that PASTA retains a normalized performance of 0.879 and 100\% success rate (without distractor: 0.896/100\%) using the same amount of samples to plan. Our planner is able to ignore the distractors using our attention structure at every step to only attend to the relevant components in the scene. We also added an example trajectory with distractor dough pieces to our website under ``CRS with distractors''.
}

\begin{table*}[ht]
\centering
\begin{tabular}{@{}lcccc}
\toprule
Planning Horizon & 3 & 4 & 5 & 6  \\
\midrule
    \hspace{5mm} Performance &
    0.896 & 0.866 & 0.90 & 0.878 \\
    \hspace{5mm} Success Rate & 5/5 & 4/5 & 5/5  & 4/5 \\
\bottomrule
\end{tabular}
\caption{\algo's performance v.s. varying the planning horizon in CRS.}
\label{tab:vary-horizon}
\end{table*}

\rebuttal{\section{Further Discussion on Limitations and Future Work}}
\label{sec:Further Discussion on Limitations and Future Work}

\rebuttal{\textbf{More Efficient Planning} Planning skill sequences with a large search space is a challenging problem by itself but much progress has been made by the task and motion planning community to obtain a plan skeleton~\cite{garrett2017sample,kim2020learning,driess2020deep}. For example, Caelan et al.~\cite{garrett2017sample} propose two methods, the first one is to interleave searching the skill sequence with lower-level optimization and the second one is to have lazy placeholders for some skills. Danny et al.~\cite{driess2020deep} propose to predict skill sequences from visual observation. Recent works have also explored finding skill sequences using pre-trained language models~\cite{ahn2022can,huang2022language}.}

\rebuttal{\textbf{Sim2Real Transfer for Real Dough} One possible approach is to train with domain randomization to make the policy more robust to changing dynamics (e.g. stickiness) of dough. Another option is to perform online system identification of the dough dynamics parameters~\cite{yu2017preparing,kumar2021rma} or real2sim methods~\cite{ramos2019bayessim,chebotar2019closing}. In future work, we can also integrate our method with other works that perform low-level dough manipulation in the real world, such as recent work from Qi et al.~\cite{qi2022learning}.}

\rebuttal{\textbf{Goal Specification} Our planner requires specifying the goal with a point cloud and coming up with a point cloud goal is not always easy. However, rapid progress is being made with language-conditioned manipulation and future work can combine language to specify more diverse tasks.}

\begin{figure}[ht]
    \centering
    \includegraphics[width=\textwidth]{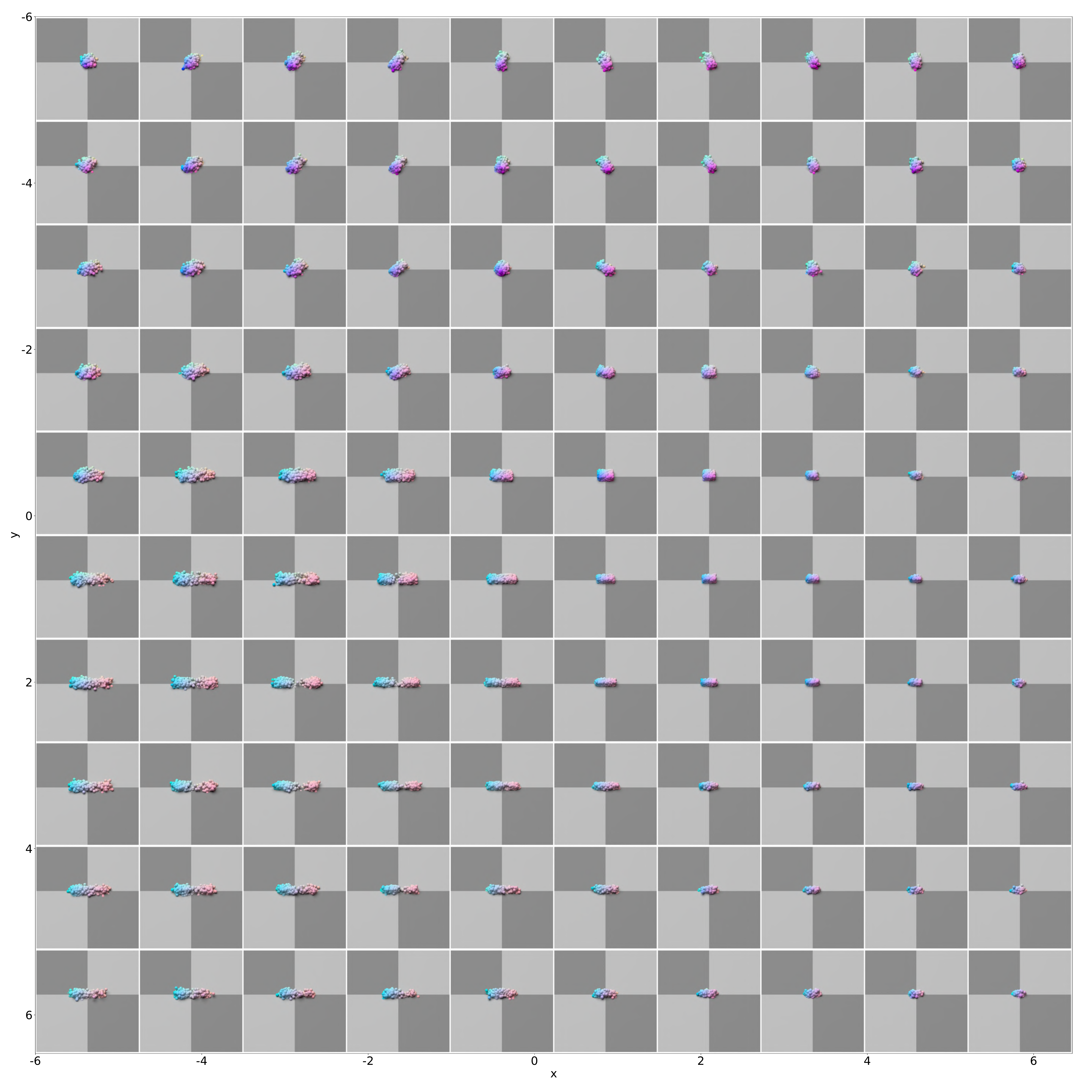}
    \caption{Latent space of PASTA in CutRearrange. We sample coordinates on a grid from the 2D latent space encoding the shapes and then decoding each latent vector into a point cloud. We then rearrange the decoded point cloud into the grid based on the corresponding coordinates in the latent space. }
    \label{fig:latent1}
\end{figure}
\begin{figure}[ht]
    \centering
    \includegraphics[width=\textwidth]{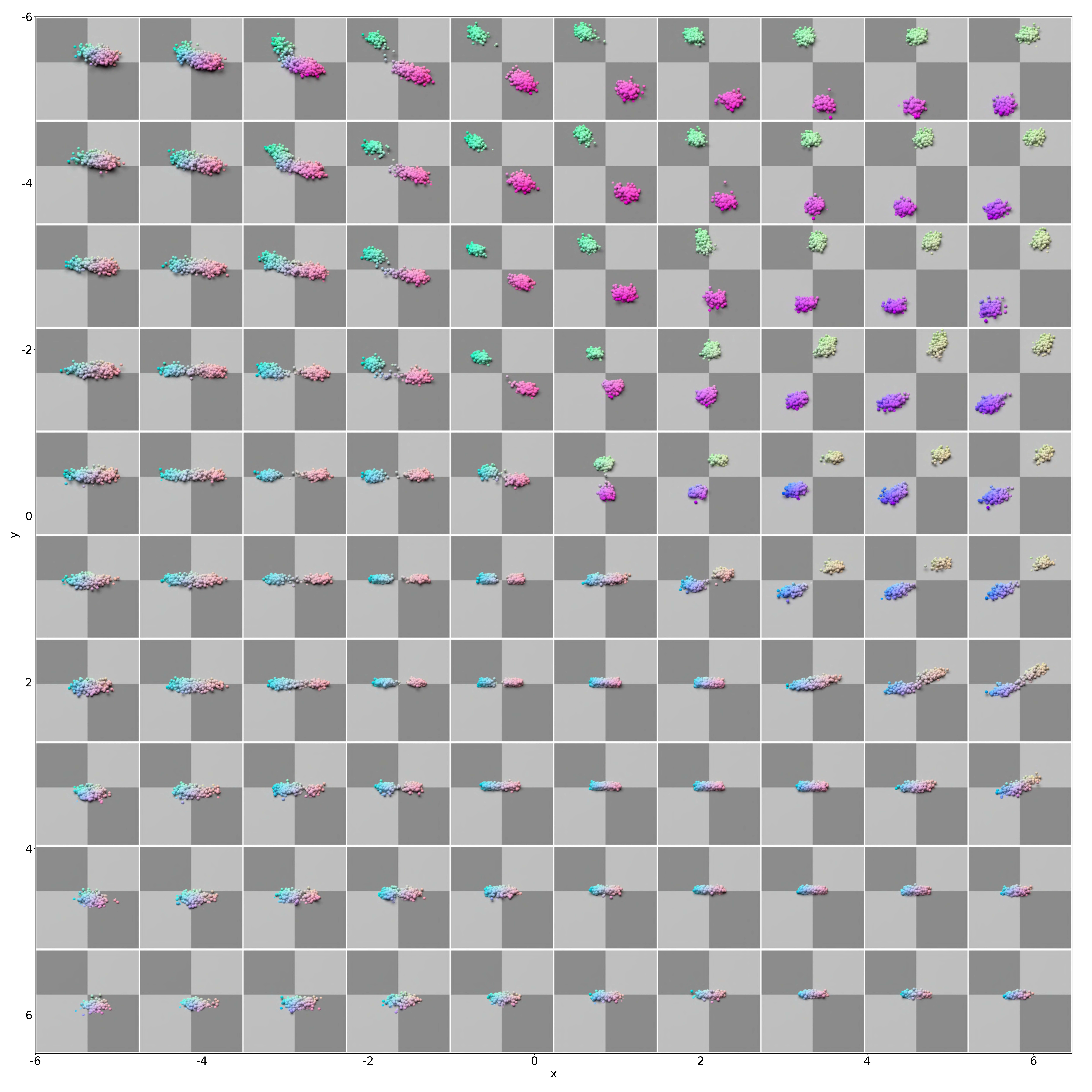}
    \caption{Latent space of Flat 3D in CutRearrange. We sample coordinates on a grid from the 2D latent space encoding the shapes and then decoding each latent vector into a point cloud. We then rearrange the decoded point cloud into the grid based on the corresponding coordinates in the latent space. }
    \label{fig:latent5}
\end{figure}
\clearpage
\newpage
\bibliography{ref}  